\documentclass[conference]{IEEEtran}
\makeatletter\if@twocolumn\PassOptionsToPackage{switch}{lineno}\else\fi\makeatother
\usepackage{bigdelim}

\usepackage{graphicx}
\usepackage[T1]{fontenc}
\usepackage[numbers,sort&compress]{natbib}
\usepackage{mathtools}
\usepackage{booktabs}
\usepackage{lineno,hyperref}
\usepackage{amsmath}
\usepackage{amsfonts}
\usepackage{amssymb}
\usepackage{amsmath}
\usepackage{multirow,morefloats,floatflt,cancel,tfrupee}
\usepackage{makecell}
\usepackage{tikz}
\usepackage{setspace}
\usepackage{color}
\usepackage{multirow}
\usepackage{adjustbox}
\usepackage{bbding}
\usepackage{physics}
\usepackage{colortbl}
\usepackage{xcolor}
\usepackage{pifont}
\usepackage[nointegrals]{wasysym}
\renewcommand\IEEEkeywordsname{Keywords}

\newcommand{\teblebold}[1]{\small\textbf{#1}}

\makeatletter

\let\old@ps@IEEEtitlepagestyle\ps@IEEEtitlepagestyle
\def\confheader#1{%
    \def\ps@IEEEtitlepagestyle{%
        \old@ps@IEEEtitlepagestyle%
        \def\@oddhead{\strut\hfill#1\hfill\strut}%
        \def\@evenhead{\strut\hfill#1\hfill\strut}%
    }%
    \ps@headings%
}
\makeatother

\begin{document}

        \title{Revolutionizing Traffic Sign Recognition: Unveiling the Potential of Vision Transformers}

\author{\IEEEauthorblockN{Susano Mingwin,
Yulong Shisu,
Yongshuai Wanwag, Sunshin Huing}\\
University of Chinese Academy of Sciences. Northwestern University}

\maketitle

\begin{abstract}

This research introduces an innovative method for Traffic Sign Recognition (TSR) by leveraging deep learning techniques, with a particular emphasis on Vision Transformers. TSR holds a vital role in advancing driver assistance systems and autonomous vehicles. Traditional TSR approaches, reliant on manual feature extraction, have proven to be labor-intensive and costly. Moreover, methods based on shape and color have inherent limitations, including susceptibility to various factors and changes in lighting conditions.
This study explores three variants of Vision Transformers (PVT, TNT, LNL) and six convolutional neural networks (AlexNet, ResNet, VGG16, MobileNet, EfficientNet, GoogleNet) as baseline models. To address the shortcomings of traditional methods, a novel pyramid EATFormer backbone is proposed, amalgamating Evolutionary Algorithms (EAs) with the Transformer architecture.
The introduced EA-based Transformer block captures multi-scale, interactive, and individual information through its components: Feed-Forward Network, Global and Local Interaction, and Multi-Scale Region Aggregation modules. Furthermore, a Modulated Deformable MSA module is introduced to dynamically model irregular locations.
Experimental evaluations on the GTSRB and BelgiumTS datasets demonstrate the efficacy of the proposed approach in enhancing both prediction speed and accuracy. This study concludes that Vision Transformers hold significant promise in traffic sign classification and contributes a fresh algorithmic framework for TSR. These findings set the stage for the development of precise and dependable TSR algorithms, benefiting driver assistance systems and autonomous vehicles.
\end{abstract}


\section{Introduction}

Driver assistance systems and driverless cars rely significantly on traffic sign recognition and detection (TSRD). A critical condition for widespread TSRD use is the development of an algorithm that is both dependable and highly accurate over a large range of real-world events. Furthermore, photographs of traffic collected along highways are often unsatisfactory, which is exacerbated by the wide range of traffic signs that must be recognized. These photographs frequently suffer from distortions produced by camera motion, inclement weather, and poor illumination, all of which considerably exacerbate the difficulties connected with this undertaking.

This field of research remains complex, with several studies undertaken throughout the years. A full collection of these research is offered in \cite{sanyal2020traffic}. TSRD consists of two independent tasks: traffic sign recognition (TSR) and traffic sign detection (TSD). Historically, TSR approaches processed dataset information mostly through manual feature extraction. Manual methods were used to extract features based on color, geometric shape, and edge detection. Color-based techniques usually include segmenting traffic sign areas within certain color spaces, such as Hue-Saturation-Intensity (HSI), Hue-Chroma-Luminance (HCL), and so on. While these systems required significant effort to manually curate big picture databases, they proved to be expensive and time-consuming. Shape- and color-based approaches were also widely used, although they had intrinsic flaws such as susceptibility to changes in lighting, occlusions, scale variations, rotations, and translations. Although machine learning can help with some of these challenges, it requires substantial annotated datasets. In recent years, deep learning has developed as a powerful solution to traffic sign identification, reaching cutting-edge results. ResNet, VGG16, and MobileNet are among the models developed as a result of deep learning research. The use of available datasets in previous research, such as the Belgium Traffic Sign Dataset (BelgiumTS)~\cite{BelgiumTS}, is also a crucial contributor to deep learning performance.

This research project focuses on approaches for traffic sign categorization, with the goal of exploring unknown terrain in the subject. To the best of the authors' knowledge, no previous study has used the Transform model directly to the job of traffic sign categorization. The Transformer's attention mechanism is a critical component of its architecture, which was initially designed to address sequence-to-sequence complications in natural language processing. Pure Transformer focuses entirely on attention operations, with no recursion or convolution. This purposeful design decision maximizes parallel efficiency while maintaining performance criteria. Although traditional recurrent neural networks (RNNs) such as GRU or LSTM, along with convolutional neural networks (CNNs), primarily concentrate on local and recurrent information, transformers focus on more global information due to their attention mechanism. Vision Transformers are cutting-edge computer vision solutions with real-world applications in autonomous driving, object detection \cite{pyramid}, healthcare \cite{Dilated, medvit}, defense \cite{Embeded}, and more (Zhang et al., 2022). This study uses three types of Vision Transformers: the Pyramid Vision Transformer model (PVT)\cite{PVT}, Transformer in Transformer (TNT)\cite{TNT}, and Locality iN Locality (LNL)\cite{LNL}. To ensure comprehensive evaluation and comparison across methodologies, six well-established convolutional neural networks (AlexNet\cite{AlexNet}, ResNet~\cite{Resnet}, VGG16~\cite{VGG16}, MobileNet~\cite{Mobilenets}, EfficientNet~\cite{efficientnet}, and GoogleNet~\cite{GoogleNet}) are used as baseline models.

This work provides a justification for the use of Vision Transformer by drawing comparisons to the well-known Evolutionary Algorithm (EA), stressing their shared mathematical basis. Taking advantage of the usefulness of various EA variations, we provide a pioneering pyramid EATFormer architecture that incorporates the suggested EA-based Transformer (EAT) block. This EAT block consists of three improved modules: the Feed-Forward Network (FFN), Global and Local Interaction (GLI), and Multi-Scale Region Aggregation (MSRA). These components are precisely designed to individually capture interactive, individual, and multi-scale information, hence improving overall performance.
Furthermore, as part of the transformer's backbone, we design a Task-Related Head (TRH) to enable flexible information fusion at the end of the process. In addition, we provide a Modulated Deformable MSA (MD-MSA) module, which allows for dynamic modeling of irregular sites. Our experimental results on the GTSRB~\cite{GTSRB} and BelgiumTS~\cite{BelgiumTS} datasets demonstrate a significant improvement in prediction speed and accuracy using our technique.

 \begin{figure*}[!t]
 \centering
  \includegraphics[width=\textwidth]{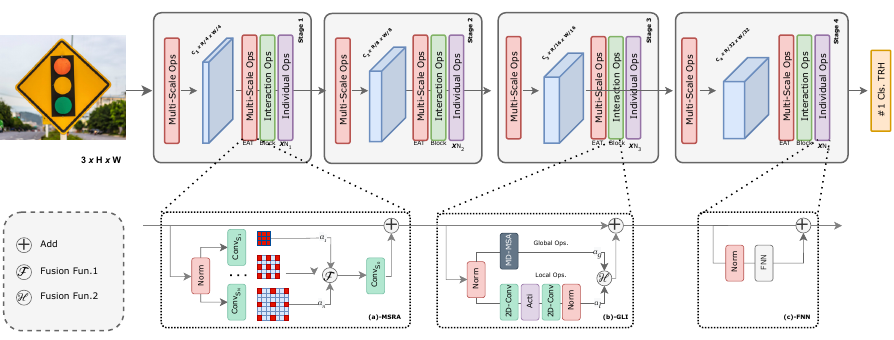}
  \caption{Network architecture of the proposed Transformer.}\label{fig.architecure}
\end{figure*}

\section{The Proposed Approach}

Figure 1 shows the upgraded EATFormer design, which consists of four stages with various resolutions based on the pyramid model~\cite{PVT}. It combines EAT blocks, each with three coupled paradigm $y=f(x)+x$ residuals: (a) FFN, (b)GLI, and (c) MSRA modules. The MSRA carries downsample process from upper stage to down stage by using convolution layer with compatible stride. In addition, we develop a novel Modulated Deformable MSA (MDMSA) module to improve global modeling and a Task-Related Head (TRH) that provides a more refined and adaptable approach to categorization.

\subsection{Overview of Vision Transformer}

In the Vision Transformer, Multi-Head Attention plays a pivotal role in capturing complex relationships between different image patches. This mechanism allows the model to focus on different parts of the input image simultaneously, facilitating parallel processing of information.
In Multi-Head Attention, the input embeddings are first transformed into queries, keys, and values using learned linear projections. These transformed representations are then split into multiple heads, each of which undergoes a separate attention computation. Within each head, attention scores are computed between the queries and keys, determining the relevance of each patch to every other patch. The resulting attention weights are then used to linearly combine the values, generating context-aware representations for each patch. Finally, the outputs from all heads are concatenated and linearly projected to produce the final Multi-Head Attention output.

The Feed-Forward Network serves as the primary mechanism for modeling complex non-linear relationships within the Vision Transformer. It consists of two linear transformations separated by a non-linear activation function, typically a GELU (Gaussian Error Linear Unit) or ReLU (Rectified Linear Unit).
In the Vision Transformer, the FFN operates independently on each position (or patch) in the input sequence. It projects the input embeddings into a higher-dimensional space through the first linear transformation, enabling the model to capture intricate features and patterns. Subsequently, a non-linear activation function introduces non-linearity into the network, enhancing its expressive power. Finally, the output is projected back to the original embedding dimension through the second linear transformation. The FFN's ability to model complex interactions between patches, coupled with its parameter-efficient design, contributes significantly to the Vision Transformer's impressive performance in various vision tasks.

By integrating Multi-Head Attention and Feed-Forward Networks as core components, the Vision Transformer achieves unparalleled performance in computer vision tasks, surpassing the traditional limitations of convolutional neural networks and unlocking new possibilities in image understanding and analysis.. Within the Transformer encoder, a series of inputs $u$ is entered, and the outputs $v$ are calculated as follows:

\begin{equation}\label{eq.1}
 \hat{u} = u + \operatorname{MSA}(\operatorname{Norm}(u))
\end{equation}

\begin{equation}\label{eq.2}
 v = \hat{u} + \operatorname{MLP}(\operatorname{Norm}((\hat{u}))
\end{equation}

\noindent where $\operatorname{Norm}$ is the normalization of layers, $\operatorname{MLP}$ is the multilayer perceptron, and $\operatorname{MSA}$ is the multi-head self-attention. After the transformer encoder, ViT utilizes the CLS for the final prediction phase.

\subsection{Multi-Scale Region Aggregation}
We expand the notion of using numerous populations and various search locations to improve model performance to 2D image analysis, inspired by EA techniques (\cite{li2021differential, zhang2022eatformer}). As part of our work on vision transformers, we offer a new module called Multi-Scale Region Aggregation (MSRA). Figure 4.(a) depicts MSRA's architecture, which integrates $N$ local convolution operations (Conv $_{S_n}, 1 \leq n \leq N$) with variable strides. By combining information from several receptive fields, these techniques successfully impose an inductive bias without requiring extra position embedding methods. Specifically, the $n$-th dilation operation $o_n$, which alters the input feature map $x$, may be expressed mathematically as follows:

$$
\begin{aligned}
& o_n(\boldsymbol{x})=\operatorname{Conv}_{S_n}(\operatorname{Norm}(\boldsymbol{x})) \\
& \text { s.t. } n=1,2, \ldots, N,
\end{aligned}
$$

To mix all operations proportionately, we introduce the Weighted Operation Mixing (WOM) mechanism, applying a softmax function over a set of trainable weights $\alpha_1, \ldots, \alpha_N$. Using the mixing function $\mathcal{F}$, the intermediate representation $\boldsymbol{x}_o$ may be obtained as follows:

$$
x_o=\sum_{n=1}^N \frac{\exp \left(\alpha_n\right)}{\sum_{n^{\prime}=1}^N \exp \left(\alpha_{n^{\prime}}\right)} o_n(\boldsymbol{x}) \text {, }
$$

In the given equation, $F$ represents the addition function. While alternate fusion functions such as concatenation may provide better results, they often need more arguments. For the purpose of simplicity, this article defaults to using the addition function. Next, a convolution layer ($Conv_{So}$) is used to transfer the intermediate representation $x_o$ to the same number of channels as the input $x$. The module's final output is obtained via relative connections. The MSRA module not only improves the regularity and beauty of the EATFormer, but it also functions as an embedding patch and stem. Because of its CNN-based MSRA, this research does not use any sort of position embedding, as the GLI module already supplies a natural inductive bias.

\subsection{Global and Local Interaction}

We propose a new enhancement to the Multi-Scale Attention (MSA) grounded global module, transforming it into a unique entity called the Global and Local Interaction (GLI) module. The idea behind this is to accelerate and improve the convergence to high-quality solutions by incorporating local exploration methods into Evolutionary Algorithms (EA), as previously described \cite{farzipour2023traffic}. To view a clearer illustration of this, see Figure 1-(c). The GLI module is explained in Figure 4-(b), which shows a local pathway concurrent with the global pathway. The local pipeline aims to extract discriminative insights relevant to localized settings, similar to the previously proposed concept of a localized population. On the other hand, the global route does not waver while modeling large global datasets.

At the channel tier, input properties are divided into two categories: local (blue) and global (green). To allow for more subtle interactions between features, these properties are processed separately via global and local routes. To achieve fair contributions from both trajectories, we use the Weighted Operation Mixing approach described in the next paragraph, fine-tuning the weights assigned to the local ($\alpha_l$) and global ($\alpha_g$) branches. The outputs of these diverse routes are then combined, restoring the data to its original dimensions. This combination may be viewed as a flexible plug-and-play concatenation operation designated as $H$, which improves the current transformer architecture.

Local operations can take many forms, including traditional convolutional layers, complex modules like Deformable Convolutional Networks (DCN) \cite{zhu2019deformable}, and local Multi-Scale Attention (MSA). In contrast, global operations can profit from MSA, Dynamic-MSA (D-MSA), Performer, and other such methods (Choromanski, 2020).

In this paper, we use simple convolution and MSA modules as the core components of the GLI module. This strategy assures that, despite the maintenance of global modeling prowess, localization is strengthened, as seen in Figure 1. Notably, the maximum path length between any two sites in our proposed architecture is still $O(1)$, retaining efficiency and parallelism similar to the standard Vision Transformer (ViT).

The determination of the feature segregation ratio, $p$, exerts a momentous influence on the efficiency and efficacy of the model. The model's parameters, floating-point operations (FLOPs), and accuracy levels vary contingent on the utilized ratio. To elucidate, the local path encompasses point-wise feature maps with dimensions of $R^{C\times H\times W} = R^{C\times L}$, while both pathways incorporate $k \times k$ depth-wise convolutions. Assuming the count of global channels as $C_g = p \times C$ and that of local channels as $C_l = C - C_g$.
Let us now delve into the parameter tally and computational constituents of the enhanced GLI module, delineated hereinafter:

$$
\begin{aligned}
\text { Params }= & 5 C_g^2+\left(2-2 C-K^2\right) C_g+ \\
& \left(k^2+2+C\right) C .
\end{aligned}
$$

 \begin{figure}[!t]
 \centering
  \includegraphics[width=\linewidth]{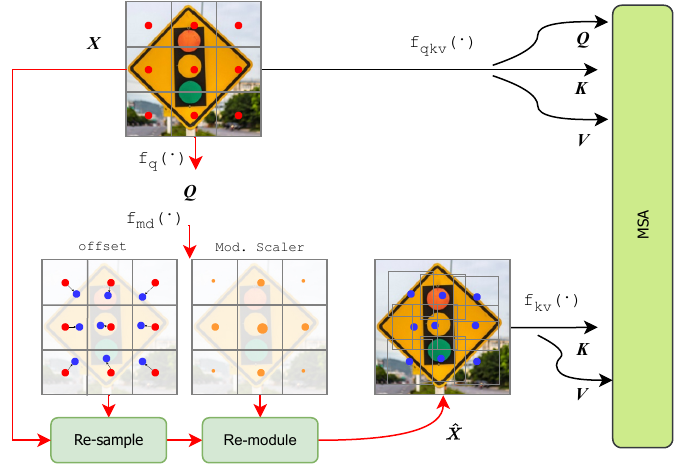}
  \caption{Structure of the proposed MD-MSA.}\label{fig.module}
  \vspace{-5mm}
\end{figure}

\subsection{Modulated Deformable MSA}

We have intricately refined a distinctive module, denoted as the Modulated Deformable Multi-Scale Attention (MD-MSA), predicated upon the intricate spatial configurations that deviate from the orthodox horizontal and vertical orientations commonly observed in real-world scenarios. This module intricately addresses the fine-tuning and re-weighting exigencies pertinent to each spatial patch. Figure 2 portrays a visual juxtaposition delineating the conventional Multi-Scale Attention (MSA) process (illustrated by the blue dotted line) vis-à-vis the MD-MSA procedure (depicted by the red solid line).
The derivation of the Query-Key-Value (QKV) features from the input feature map $X$ is orchestrated through the function $f_{qkv}(\cdot)$, expressed as $Q K V=f_{qkv}(\boldsymbol{X})$, wherein $f_{qkv}=f_q \oplus f_k \oplus f_v$ (wherein the symbol $\oplus$ denotes a concatenation operation amalgamating the functions $f_q$, $f_k$, and $f_v$). In contradistinction, the MD-MSA modus operandi introduces nuanced alterations. The proposed MD-MSA diverges from the native MSA in its utilization of the meticulously fine-tuned feature map $\hat{X}$ to elicit the $\boldsymbol{K}$ and $\boldsymbol{V}$ features in a discernibly query-centric manner.
To expatiate further, considering an input feature map $X$ encompassing $L$ loci, the function $f_q$ is harnessed to fabricate the query matrix $Q$, epitomized as $\boldsymbol{Q}=f_q(\boldsymbol{X})$. Subsequently, this query matrix is deployed to prognosticate the deformable offsets $\Delta l$ and modulation scalars $\Delta m$ for all points, thereby enabling access to the refined feature map $\Delta m$.
$$
\Delta l, \Delta m=f_{m d}(\boldsymbol{Q})
$$

Using the $l$-th position, we calculate the resampled and reweighted feature $\hat X_l$ as follows:

$$
\hat{\boldsymbol{X}}_l=\mathcal{S}\left(\boldsymbol{X}_l, \Delta l\right) \cdot \Delta m
$$

$\Delta l$ is the relative coordinate for the $l$-th position and has an unbounded range, whereas $\Delta m$ is limited to the interval $(0,1$. The symbol $\mathcal{S}$ refers to the bilinear interpolation function. As a result, the updated feature map $\operatorname{map} \hat{\boldsymbol{X}}$ yields $\boldsymbol{K} \boldsymbol{V}=f_{k v}(\hat{\boldsymbol{X}})$. MD-MSA stands out from current literature~\cite{xia2022vision} largely because to its modulation operation. The MD-MSA approach improves outcomes by carefully examining a wide range of spatial factors.

\begin{table*}[t]
    \caption{Performance comparison of SOTA models and proposed Transformer on GTSRB.}
    \begin{adjustbox}{width=.9\textwidth,center}

     \begin{tabular}{|c|c|c|c|c|c|c|c|c|}
        \toprule
         \teblebold{Method} & \teblebold{Accuracy $(\%)$} & Precison (\%) & Recall (\%) & F1-Score (\%) & $\begin{array}{l}\text { Parameters } \\
        \text { (in Millions) }\end{array}$ & $\begin{array}{l}\text { Training Time } \\
        \text { (in Seconds) }\end{array}$ & $\begin{array}{l}\text { Response Time } \\
        \text { (in millieconds) }\end{array}$ & $\begin{array}{l}\text { Resources Usage } \\
        \text { (in MByte) }\end{array}$ \\

        \midrule AlexNet & 97.08 & 97.17 & 97.08 & 97.07 & 15.78 & 1638.5 & 73.05 & 28.0 \\
        \hline GoogleNet & 96.97 & 97.09 & 96.97 & 96.92 & 13.70 & 2983.7 & 94.80 & 22.3 \\
        \hline EfficientNet & 97.44 & 97.82 & 97.94 & 97.63 & 43.80 & 2454.8 & 90.68 & 23.7 \\
        \hline MobileNetV2 & 97.30 & 97.45 & 97.30 & 97.30 & 12.77 & 1319.7 & 100.17 & 23.0 \\
        \hline VGG16 & 97.60 & 97.67 & 97.60 & 97.59 & 22.40 & 2442.9 & 96.21 & 24.1 \\
        \hline ResNet50V2 & 98.31 & 98.37 & 98.31 & 98.27 & 40.37 & 1974.9 & 105.93 & 26.5 \\
        \hline TNT & 97.17 & 97.44 & 97.32 & 97.38 & 23.47 & 1974.9 & 105.93 & 26.5 \\
        \hline LNL & 98.29 & 98.18 & 98.37 & 98.23 & 23.80 & 1974.9 & 105.93 & 26.5 \\
        \hline PVT & 96.83 & 96.79 & 96.74 & 96.92 & 24.56 & 1974.9 & 105.93 & 26.5 \\
        \hline Ours & \teblebold{98.41} & \teblebold{98.51} & \teblebold{98.41} & \teblebold{98.42} & \teblebold{9.61} & \teblebold{1264.2} & \teblebold{74.34} & \teblebold{20.7} \\
        \bottomrule
        \end{tabular}
    \end{adjustbox}
\end{table*}

\section{Experiments}

This section begins by discussing the datasets used to train our revised Transformer model and provides an overview of the experimental configurations. The experimental findings are then thoroughly examined, including a comparison of our Transformer model's performance to state-of-the-art research in traffic sign identification.

\subsection{Datasets and Implementation details}

This study evaluates the model's effectiveness using two publicly available datasets: GTSRB~\cite{GTSRB} and BelgiumTS~\cite{BelgiumTS}. These datasets, which are well-known as standards for multiclass traffic sign classification, include a wide range of complex and sophisticated traffic signs. They accurately depict real-world scenes by including elements like as tilt, uneven illumination, distortion, occlusion, and comparable backdrop hues. The model is rigorously evaluated using a large sample of complicated and difficult-to-distinguish traffic indicators from various datasets.

Furthermore, our suggested model is especially designed for deployment in real-world circumstances where safety is of the utmost importance, with a focus on accuracy and dependability. We chose a picture resolution of 224 × 224 to balance classification accuracy and model complexity. Furthermore, we rigorously train and assess our model using the GTSRB and BelgiumTS datasets to ensure its efficacy and dependability for traffic sign identification.
Throughout the training procedure, we used the Adam optimizer with a learning rate of 0.008 and a batch size of 50. We use data augmentation techniques like random rotations, zooming, and horizontal flips to increase the size of the training set and improve the model's generalization capabilities.
The tests are carried out using Pytorch library with a Titan RTX 24G, and the major code implementation is in Python 3.8.

\begin{table}[t]
    \caption{Performance for Traffic Sign Recognition on the BelgiumTS.}
    \begin{adjustbox}{width=.95\linewidth,center}

    \begin{tabular}{lccccc}
        \hline Method & Accuracy (\%) & Precison (\%) & Recall (\%) & F1-Score (\%) & $\begin{array}{c}\text { Parameters } \\
        \text { (in Millions) }\end{array}$ \\

        \hline AlexNet & 70.71 & 80.77 & 70.71 & 72.02 & 15.79 \\
        \hline GoogleNet & 12.46 & 40.15 & 12.46 & 12.86 & 13.71 \\
        \hline EfficientNet & 84.08 & 88.47 & 84.08 & 84.74 & 43.80 \\
        \hline MobileNetV2 & 36.98 & 71.89 & 36.98 & 35.75 & 12.77 \\
        \hline VGG16 & 69.12 & 79.09 & 69.12 & 69.82 & 22.41 \\
        \hline ResNet50V2 & 84.20 & 89.82 & 84.20 & 85.46 & 40.37 \\
        \hline TNT & 83.15 & 88.52 & 83.50 & 84.92 & 23.47 \\
        \hline LNL & 84.65 & 89.15 & 84.79 & 87.11 & 23.80 \\
        \hline PVT & 83.04 & 87.81 & 83.06 & 83.29 & 24.56 \\
        \hline Ours & $\mathbf{9 2 . 1 6}$ & $\mathbf{9 4 . 83}$ & $\mathbf{9 2 . 17}$ & $\mathbf{9 2 . 5 4}$ & $\mathbf{9 . 6 3}$ \\
        \hline
        \end{tabular}
    \end{adjustbox}
\end{table}

\subsection{Evaluation Measures}

F1-score, accuracy, recall, and precision stand as customary pivotal metrics for evaluating the prowess of classification algorithms. The accuracy gauge quantifies the overarching correctness of the model's prognostications, derived by dividing the count of accurately classified instances by the aggregate number of instances, as delineated in Equation (3). Precision gauges the proportion of true positives amidst all positive prognostications proffered by the model, formulated per Equation (4). Meanwhile, recall computes the ratio of authentic positives amidst all factual positive instances within the dataset, explicated via Equation (5). The f1-score, an amalgam of accuracy and recall, furnishes a nuanced evaluation amalgamating these two indices into a balanced measure, as expounded in Equation (6).
Throughout our experimental endeavors, we shall employ these cardinal metrics to gauge the efficacy of our proposed methodology. Through the lens of these gauges, we can scrutize the model's adeptness in accurately discerning traffic signs.

\begin{equation}
A C C =\frac{T_P+T_N}{T_P+T_N+F_P+F_N}\\
\end{equation}
\begin{equation}
\text { recall } =\frac{T_P}{T_P+F_N} \\
\end{equation}
\begin{equation}
\text { precision } =\frac{T_P}{T_P+F_P} \\
\end{equation}
\begin{equation}
F_{1}-\text { score } =\frac{2 \times \text { precision } \times \text { recall }}{\text { precision }+\text { recall }}
\end{equation}

\subsection{Comparison with the state-of-the-art}

\textbf{GTSRB} Evaluation: To evaluate the usefulness and reliability of our suggested Transformer model for Traffic Sign Recognition, we did extensive tests on the GTSRB dataset. To establish a benchmark, we compared our model's results with several state-of-the-art techniques, encompassing Transformer variants (PVT~\cite{PVT}, TNT~\cite{TNT}, LNL~\cite{LNL}) and CNN methods (AlexNet~\cite{AlexNet}, ResNet~\cite{Resnet}, VGG16~\cite{VGG16}, MobileNet~\cite{Mobilenets}, EfficientNet~\cite{efficientnet}, GoogleNet~\cite{GoogleNet}).The experimental results clearly indicate that our proposed model performed very well, highlighting its potential for use in real-time traffic safety and driver aid systems. Table 1 provides a detailed comparison of our model to various cutting-edge models using the GTSRB dataset. Our suggested model outperformed AlexNet, ResNet, VGG16, EfficientNet, GoogleNet, PVT, and LNL, with gains in accuracy of 1.33\%, 0.1\%, 0.81\%, 0.97\%, 1.44\%, 1.58\%, and 0.12\%, respectively. It is worth noting that the suggested model excels in computing efficiency and has a small model size, making it extremely practical and appropriate for real-time TSR applications.

\textbf{BelgiumTS:} To evaluate the proposed TSR model's efficacy and robustness, it was compared to well-known models such as PVT~\cite{PVT}, TNT~\cite{TNT}, AlexNet~\cite{AlexNet}, and GoogleNet~\cite{GoogleNet} on the BelgiumTS~\cite{BelgiumTS} dataset. The BelgiumTS dataset, which contains 4785 training photos and 2250 testing images, is restricted in size, therefore we developed synthetic images and expanded the dataset using augmentation techniques such as horizontal flips, magnification, and random rotations. This augmentation strategy allowed the model to learn from a broader set of pictures, lowering generalization mistakes and boosting overall performance.

The results in Table 2 demonstrate that the suggested model outperforms other techniques in terms of classification performance. The findings clearly show that the proposed model outperforms all others on the BelgiumTS dataset while requiring the fewest parameters. The results show that the system outperforms AlexNet, ResNet, VGG16, MobileNet, EfficientNet, and GoogleNet in terms of F1-score, accuracy, recall, and precision.
Specifically, the suggested model obtains an accuracy rate of 92.16\%, outperforming AlexNet by 21.45 percentage points, EfficientNet by 8.08 percentage points, TNT by 9.01 percentage points, and LNL by 7.51 percent. Furthermore, the suggested model achieves a precision of 94.83\%, beating ResNet and LNL with precisions of 89.82\% and 89.15\%, respectively. Furthermore, with a recall score of 92.17\%, the suggested model outperforms TNT (84.92\%) and LNL (87.11\%).

Furthermore, a comparison of our suggested Transformer with existing techniques and models reveals significant gains in terms of model size, accuracy, and computing efficiency. These qualities make it ideal for real-time TSR applications. Our method's practical usefulness is further enhanced by its ability to improve road safety through fast and exact recognition and detection of traffic signs.

\section{Conclusion}

The development of reliable and extremely precise algorithms for traffic sign identification and detection is critical for the broad integration of driver assistance systems and self-driving cars. Numerous studies have been conducted in this area over the years, with traditional approaches focusing on manual feature extraction, which has proven to be both expensive and time-consuming. However, the introduction of deep learning, particularly the Transformer architecture, has resulted in a paradigm change in traffic sign identification, with state-of-the-art performance.This work describes a novel pyramid EATFormer backbone that incorporates an Evolutionary Algorithm (EA)-based Transformer (EAT) block and demonstrates its effectiveness in improving prediction speed and accuracy on real-world datasets. The findings highlight Vision Transformers' potential as a leading alternative for traffic sign categorization in practical applications.

\bibliographystyle{IEEEtran}

\bibliography{article}

\begin{thebibliography}{10}
\providecommand{\url}[1]{#1}
\csname url@samestyle\endcsname
\providecommand{\newblock}{\relax}
\providecommand{\bibinfo}[2]{#2}
\providecommand{\BIBentrySTDinterwordspacing}{\spaceskip=0pt\relax}
\providecommand{\BIBentryALTinterwordstretchfactor}{4}
\providecommand{\BIBentryALTinterwordspacing}{\spaceskip=\fontdimen2\font plus
\BIBentryALTinterwordstretchfactor\fontdimen3\font minus
  \fontdimen4\font\relax}
\providecommand{\BIBforeignlanguage}[2]{{%
\expandafter\ifx\csname l@#1\endcsname\relax
\typeout{** WARNING: IEEEtran.bst: No hyphenation pattern has been}%
\typeout{** loaded for the language `#1'. Using the pattern for}%
\typeout{** the default language instead.}%
\else
\language=\csname l@#1\endcsname
\fi
#2}}
\providecommand{\BIBdecl}{\relax}
\BIBdecl

\bibitem{sanyal2020traffic}
B.~Sanyal, R.~K. Mohapatra, and R.~Dash, ``Traffic sign recognition: A
  survey,'' in \emph{2020 International Conference on Artificial Intelligence
  and Signal Processing (AISP)}.\hskip 1em plus 0.5em minus 0.4em\relax IEEE,
  2020, pp. 1--6.

\bibitem{BelgiumTS}
M.~Mathias, R.~Timofte, R.~Benenson, and L.~Van~Gool, ``Traffic sign
  recognition—how far are we from the solution?'' in \emph{The 2013
  international joint conference on Neural networks (IJCNN)}.\hskip 1em plus
  0.5em minus 0.4em\relax IEEE, 2013, pp. 1--8.

\bibitem{pyramid}
O.~N. Manzari, A.~Boudesh, and S.~B. Shokouhi, ``Pyramid transformer for
  traffic sign detection,'' in \emph{2022 12th International Conference on
  Computer and Knowledge Engineering (ICCKE)}.\hskip 1em plus 0.5em minus
  0.4em\relax IEEE, 2022, pp. 112--116.

\bibitem{Dilated}
D.~Saadati, O.~N. Manzari, and S.~Mirzakuchaki, ``Dilated-unet: A fast and
  accurate medical image segmentation approach using a dilated transformer and
  u-net architecture,'' \emph{arXiv preprint arXiv:2304.11450}, 2023.

\bibitem{medvit}
O.~N. Manzari, H.~Ahmadabadi, H.~Kashiani, S.~B. Shokouhi, and A.~Ayatollahi,
  ``Medvit: A robust vision transformer for generalized medical image
  classification,'' \emph{Computers in Biology and Medicine}, vol. 157, p.
  106791, 2023.

\bibitem{Embeded}
O.~N. Manzari and S.~B. Shokouhi, ``A robust network for embedded traffic sign
  recognition,'' in \emph{2021 11th International Conference on Computer
  Engineering and Knowledge (ICCKE)}.\hskip 1em plus 0.5em minus 0.4em\relax
  IEEE, 2021, pp. 447--451.

\bibitem{PVT}
W.~Wang, E.~Xie, X.~Li, D.-P. Fan, K.~Song, D.~Liang, T.~Lu, P.~Luo, and
  L.~Shao, ``Pyramid vision transformer: A versatile backbone for dense
  prediction without convolutions,'' in \emph{Proceedings of the IEEE/CVF
  international conference on computer vision}, 2021, pp. 568--578.

\bibitem{TNT}
K.~Han, A.~Xiao, E.~Wu, J.~Guo, C.~Xu, and Y.~Wang, ``Transformer in
  transformer,'' \emph{Advances in Neural Information Processing Systems},
  vol.~34, pp. 15\,908--15\,919, 2021.

\bibitem{LNL}
O.~N. Manzari, H.~Kashiani, H.~A. Dehkordi, and S.~B. Shokouhi, ``Robust
  transformer with locality inductive bias and feature normalization,''
  \emph{Engineering Science and Technology, an International Journal}, vol.~38,
  p. 101320, 2023.

\bibitem{AlexNet}
A.~Krizhevsky, I.~Sutskever, and G.~E. Hinton, ``Imagenet classification with
  deep convolutional neural networks,'' \emph{Communications of the ACM},
  vol.~60, no.~6, pp. 84--90, 2017.

\bibitem{Resnet}
M.~Rahimzadeh and A.~Attar, ``A modified deep convolutional neural network for
  detecting covid-19 and pneumonia from chest x-ray images based on the
  concatenation of xception and resnet50v2,'' \emph{Informatics in medicine
  unlocked}, vol.~19, p. 100360, 2020.

\bibitem{VGG16}
K.~Simonyan and A.~Zisserman, ``Very deep convolutional networks for
  large-scale image recognition,'' \emph{arXiv preprint arXiv:1409.1556}, 2014.

\bibitem{Mobilenets}
A.~G. Howard, M.~Zhu, B.~Chen, D.~Kalenichenko, W.~Wang, T.~Weyand,
  M.~Andreetto, and H.~Adam, ``Mobilenets: Efficient convolutional neural
  networks for mobile vision applications,'' \emph{arXiv preprint
  arXiv:1704.04861}, 2017.

\bibitem{efficientnet}
M.~Tan and Q.~Le, ``Efficientnet: Rethinking model scaling for convolutional
  neural networks,'' in \emph{International conference on machine
  learning}.\hskip 1em plus 0.5em minus 0.4em\relax PMLR, 2019, pp. 6105--6114.

\bibitem{GoogleNet}
C.~Szegedy, W.~Liu, Y.~Jia, P.~Sermanet, S.~Reed, D.~Anguelov, D.~Erhan,
  V.~Vanhoucke, and A.~Rabinovich, ``Going deeper with convolutions,'' in
  \emph{Proceedings of the IEEE conference on computer vision and pattern
  recognition}, 2015, pp. 1--9.

\bibitem{GTSRB}
J.~Stallkamp, M.~Schlipsing, J.~Salmen, and C.~Igel, ``The german traffic sign
  recognition benchmark: a multi-class classification competition,'' in
  \emph{The 2011 international joint conference on neural networks}.\hskip 1em
  plus 0.5em minus 0.4em\relax IEEE, 2011, pp. 1453--1460.

\bibitem{li2021differential}
X.~Li, L.~Wang, Q.~Jiang, and N.~Li, ``Differential evolution algorithm with
  multi-population cooperation and multi-strategy integration,''
  \emph{Neurocomputing}, vol. 421, pp. 285--302, 2021.

\bibitem{zhang2022eatformer}
J.~Zhang, X.~Li, Y.~Wang, C.~Wang, Y.~Yang, Y.~Liu, and D.~Tao, ``Eatformer:
  improving vision transformer inspired by evolutionary algorithm,''
  \emph{arXiv preprint arXiv:2206.09325}, 2022.

\bibitem{farzipour2023traffic}
A.~Farzipour, O.~N. Manzari, and S.~B. Shokouhi, ``Traffic sign recognition
  using local vision transformer,'' in \emph{2023 13th International Conference
  on Computer and Knowledge Engineering (ICCKE)}.\hskip 1em plus 0.5em minus
  0.4em\relax IEEE, 2023, pp. 191--196.

\bibitem{zhu2019deformable}
X.~Zhu, H.~Hu, S.~Lin, and J.~Dai, ``Deformable convnets v2: More deformable,
  better results,'' in \emph{Proceedings of the IEEE/CVF conference on computer
  vision and pattern recognition}, 2019, pp. 9308--9316.

\bibitem{xia2022vision}
Z.~Xia, X.~Pan, S.~Song, L.~E. Li, and G.~Huang, ``Vision transformer with
  deformable attention,'' in \emph{Proceedings of the IEEE/CVF conference on
  computer vision and pattern recognition}, 2022, pp. 4794--4803.

\end{thebibliography}

\end{document}